\documentclass{article}
\usepackage{PRIMEarxiv}

\usepackage[colorlinks,urlcolor=blue,linkcolor=blue,citecolor=blue]{hyperref}

\usepackage{color, xcolor, soul, tikz, svg, subcaption, adjustbox, float}
\usepackage{caption, makecell, multirow, tabularx, amsmath, amssymb, tcolorbox}
\usepackage{lipsum, graphicx, verbatim, booktabs, hyperref, url, array}

\usepackage[T1]{fontenc}

\definecolor{cust-yellow}{HTML}{e78700}
\definecolor{cust-green}{HTML}{5aab31}
\definecolor{cust-blue}{HTML}{3074ae}

\definecolor{lime}{HTML}{A6CE39}
\DeclareRobustCommand{\orcidicon}{
	\begin{tikzpicture}
	\draw[lime, fill=lime] (0,0) 
	circle [radius=0.16] 
	node[white] {{\fontfamily{qag}\selectfont \tiny ID}};
	\draw[white, fill=white] (-0.0625,0.095) 
	circle [radius=0.007];
	\end{tikzpicture}
	\hspace{-2mm}
}
\foreach \x in {A, ..., Z}{\expandafter\xdef\csname orcid\x\endcsname{\noexpand\href{https://orcid.org/\csname orcidauthor\x\endcsname}
			{\noexpand\orcidicon}}
}

\title{PersoDPO: Scalable Preference Optimization for Instruction-Adherent, Persona-Grounded Dialogue via Multi-LLM Evaluation}

\author{
    Saleh Afzoon$^{1}$ \orcidA{}, 
    MohammadHossein Ahmadi$^{1}$\orcidB{}, 
    Usman Naseem$^{1}$\orcidC{},
    Amin Beheshti$^{1}$ \orcidD{}\\
    $^{1}$School of Computing, Macquarie University, Sydney, Australia \\
    \small \texttt{\{saleh.afzoon, mohammadhossein.ahmadi\}@hdr.mq.edu.au},\small \texttt{ \{usman.naseem, amin.beheshti\}@mq.edu.au}
}

\begin{document}
\maketitle              


\begin{abstract}

Personalization and contextual coherence are two essential components in building effective persona-grounded dialogue systems. These aspects play a crucial role in enhancing user engagement and ensuring responses are more relevant and consistent with user identity. However, recent studies indicate that open-source large language models (LLMs) continue to struggle to generate responses that are both contextually grounded and aligned with persona cues, despite exhibiting strong general conversational abilities like fluency and naturalness. We present PersoDPO, a scalable preference optimisation framework that uses supervision signals from automatic evaluations of responses generated by both closed-source and open-source LLMs to fine-tune dialogue models. The framework integrates evaluation metrics targeting coherence and personalization, along with a length-format compliance feature to promote instruction adherence. These signals are combined to automatically construct high-quality preference pairs without manual annotation, enabling a scalable and reproducible training pipeline. Experiments on the FoCus dataset show that an open-source language model fine-tuned with the PersoDPO framework consistently outperforms strong open-source baselines and a standard Direct Preference Optimization (DPO) variant across multiple evaluation dimensions.

\end{abstract}


\section{Introduction}

Dialogue systems have become increasingly integral to a wide range of applications, including virtual assistants, customer service chatbots, and task-oriented agents. As these systems are deployed in more complex and user-centric scenarios, their effectiveness depends not only on producing fluent responses but also on delivering meaningful, context-aware, and personalized responses \cite{chen2024recent}.
From a system-level perspective, high-quality dialogue systems must exhibit several critical properties—contextual understanding, persona alignment, and instruction adherence to ensure consistent response structure. Among these, contextualization (the ability to maintain coherence across dialogue turns) and personalization (the ability to tailor responses to individual user personas) are especially important for sustaining user satisfaction and trust. In addition, instruction adherence is crucial in automated pipelines, where consistent response formatting enables seamless downstream processing and system integration.

While recent advances in large language models (LLMs) have significantly improved dialogue fluency and naturalness, their ability to generate responses that are both contextually coherent and persona-aligned remains limited. As shown in the PersoBench benchmark \cite{afzoon2024persobench}, current mid-sized (around 7B–8B parameters) instruction-tuned LLMs perform well on surface-level linguistic features but fall short in deeper personalization and multi-turn coherence.
To address these limitations, recent works have explored preference-based modeling for improving dialogue quality. A knowledge-grounded dialogue model~\cite{wang2023knowledge} enhances knowledge selection through supervised contrastive learning, while DPOC~\cite{cheng2024dialogues} focuses on persona alignment by ranking responses from model variants using heuristic quality assessments. Although these studies target specific dimensions such as knowledge integration or persona consistency, they remain limited in scope. 

In contrast, our work addresses a broader combination of personalization, contextual coherence, and instruction adherence. By leveraging outputs from both open- and closed-source LLMs and relying on automatic evaluation metrics, our framework enables scalable multi-aspect fine-tuning.
We present PersoDPO, a scalable preference optimization framework that inherits complementary strengths from diverse LLMs to jointly enhance coherence, persona alignment, and instruction adherence in dialogue generation. By applying multi-aspect evaluation metrics to responses from diverse LLMs, PersoDPO fine-tunes a robust instruction-tuned model without relying on human-labeled supervision.
More specifically, the following are our research contributions:

\begin{enumerate}

    \item We propose PersoDPO, a preference optimization framework designed to jointly improve coherence, persona alignment, and instruction adherence in dialogue generation.
    
    \item We introduce a scalable method for generating preference pairs from open- and closed-source LLMs, enabling annotation-free supervision while capturing their complementary behaviors.
    
    \item We demonstrate that PersoDPO outperforms strong open-source baselines and a DPO-trained variant on the FoCus dataset across multiple evaluation dimensions.
    
\end{enumerate}

The implementation of our work, including evaluation results and generated response logs, is publicly available at: \href{https://github.com/salehafzoon/PersoDPO}{\texttt{github.com/salehafzoon/PersoDPO}
}

\section{Related Work}\label{sec2}

\subsection{Persona-Grounded Dialogue Generation}

Persona-grounded dialogue generation aims to produce responses that reflect a user’s traits or preferences, improving coherence and personalization \cite{afzoon2025modeling}.
Early work relied on pre-defined persona profiles. The PersonaChat dataset facilitated training on persona-description-response triplets \cite{zhang-etal-2018-personalizing}, and models like TransferTransfo \cite{wolf2019transfertransfo} fine-tuned transformers for more consistent and engaging responses. Profile Memory Networks \cite{zhang-etal-2018-personalizing} used persona facts as memory, though static profiles limited scalability.
To overcome these constraints, persona extraction methods emerged. PAED reformulates zero-shot persona extraction as structured triplet extraction, enabling more nuanced personalization \cite{zhu2023paed}. Building on this, DPOC \cite{cheng2024dialogues} segments dialogue into persona-relevant utterances and generates coherent summaries, which are fused with context using gated attention for dynamic identity modeling.

A parallel line of research focuses on integrating external knowledge and persona cues to enhance empathy and coherence in support-oriented dialogues. A dynamic knowledge filtering mechanism \cite{hao2025enhancing}, combined with persona extraction, improves emotional understanding in conversational agents. Techniques adapted from recommender systems, such as collaborative filtering, infer user norms implicitly, reducing the need for explicit persona definitions \cite{serramia2022collaborative}.
In mental health contexts, domain-specific personalization has advanced through cross-attentive fusion of persona traits and commonsense knowledge \cite{hao2025enhancing}. Context-aware summarization approaches also personalize medical dialogues by tracking psychological states across sessions \cite{liu2024context}. These shifts reflect a move from rule-based persona modeling to adaptive, emotionally intelligent systems.

\subsection{Evaluation Frameworks and Behavioral Constraints in LLMs}


\subsubsection{Core Benchmarks and Evaluation Metrics:}
Evaluating LLMs demands diverse and practical benchmarks that reflect real-world usage. Initial methods relied heavily on automatic metrics such as BLEU, ROUGE, and METEOR, which primarily measure surface-level textual similarity. However, these proved insufficient for assessing complex behaviors like coherence, helpfulness, or factual consistency. To address this, newer evaluations include human preference modeling frameworks such as Helpful and Harmless Assistant (HHA) and Chatbot Arena, which collect pairwise human judgments to assess alignment quality \cite{chiang2024chatbot}.

Importantly, PersoBench \cite{afzoon2024persobench} introduced a dedicated benchmarking framework for evaluating persona-consistent dialogue generation.
It was the first evaluation framework to systematically apply established multi-aspect metrics within a persona-aware benchmark. This framework also highlighted the trade-offs between helpfulness and identity coherence in LLM outputs \cite{yoon2024evaluating}. Tool-augmented LLM evaluations like Coherence-Aware RLHF and Diverse Dialogue Benchmarking further incorporate task relevance and dialogic flow into scoring systems \cite{lee2022improving,sun2024cebench}. In parallel, TimeCHARA \cite{ahn2024timechara} and RoleLLM \cite{wang2023rolellm} have been introduced in the role-playing domain to assess LLMs’ adherence to predefined characters or roles in character-based, scene-based, and temporal setups—broadening the scope of behavioral evaluation beyond generic instruction-following.

\subsubsection{Feedback Signals and Optimization Methods:}
A major innovation in LLM alignment has been the use of reinforcement learning from human feedback (RLHF), where models are fine-tuned using reward signals derived from user preferences. Traditional RLHF assumes a single reward model, but recent studies suggest modeling heterogeneity using personalized reward learners \cite{park2024rlhf}. Techniques like DPO and KL-regularized reinforcement learning \cite{huang2023trustgpt} offer alternatives that enhance optimization stability. Furthermore, coactive learning represents an important direction in using weak but implicit feedback—such as user edits—to iteratively refine models without needing explicit preference labeling \cite{tucker2024coactive}. Complementarily, active learning approaches using expected entropy reduction let models infer user preferences through strategic questioning, facilitating efficient adaptation \cite{piriyakulkij2023active}.

\subsection{Preference Optimization and Feedback-Based Learning}

The field of dialogue generation has increasingly embraced preference optimization and feedback-based learning to achieve fine-grained behavioral alignment and personalization \cite{afzoon2025modeling}. These paradigms aim to improve response quality by aligning model outputs with desirable traits such as coherence, persona consistency, and instruction compliance, often without explicit supervised labels.

Preference optimization has evolved significantly, with methods like Direct DPO and RLHF gaining traction. While RLHF relies on costly human annotations, DPO offers a more efficient alternative by directly optimizing models using comparative preference data. Notably, DPOC \cite{cheng2024dialogues} leverages user feedback across turns to build models that improve interaction-level coherence and personalization, highlighting the role of implicit feedback as a rich optimization signal. This aligns with the broader shift towards implicit reward modeling and scalable feedback mechanisms explored in other works \cite{chen2024recent}.

Behavioral alignment presents challenges, as models often appear fluent while lacking persona fidelity or intent understanding. Knowledge-grounded systems now leverage contrastive preference learning to distinguish between high- and low-value responses, enhancing informativeness without compromising personalization \cite{wang2023knowledge}.

Recent work in preference-based learning emphasizes automatic construction of preference pairs as a scalable alternative to manual labels. Instead of annotations, models infer preferences using learned rewards, heuristics, or implicit signals. Multi-dimensional evaluation criteria—such as coherence, user alignment, and formatting—enable more precise control over generation and reduce supervision needs \cite{cheng2024dialogues,park2024rlhf}.

\section{Methodology}\label{sec3}


\textbf{Problem Statement.} Starting with a formal task definition, we follow the formulation introduced in prior work~\cite{chen2023towards} on personalized dialogue generation. Given a dialogue context $C = \{u_1, \dots, u_m\}$ and a set of persona descriptions $P = \{p_1, \dots, p_n\}$, the goal is to generate a response $r$ that is both contextually grounded and aligned with the specified persona. The generation objective is modeled as a conditional language generation task:

\begin{equation}
P(r \mid C, P; \theta) = \prod_{t=1}^{T} P(r_t \mid r_{1:t-1}, C, P; \theta)
\end{equation}

Here, $r_t$ represents the $t$-th token of the response, and $\theta$ denotes the parameters of the underlying language model.

\subsection{PersoDPO}

\begin{figure*}[!h]
\centering
  \includegraphics[width=0.8\textwidth]{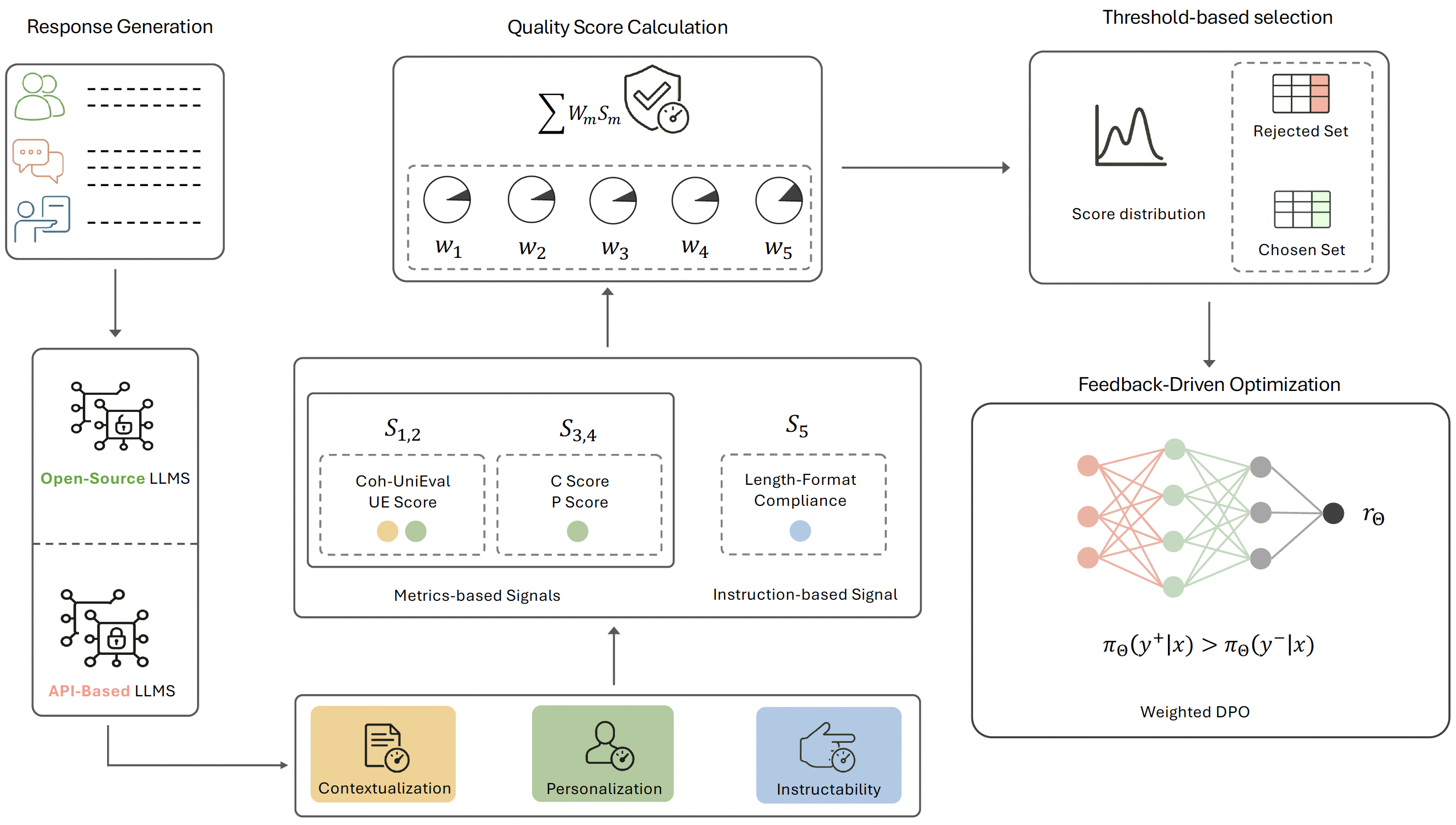}
  \caption{PersoDPO framework overview.}
  \label{fig:perso-dpo-overview}
\end{figure*}

Fig.~\ref{fig:perso-dpo-overview} illustrates the overall pipeline of the PersoDPO framework. The process begins by passing persona-aware dialogues to a diverse set of open- and closed-source LLMs within instruction-augmented prompts. This heterogeneous pool of models is selected to capture complementary strengths, reflecting variations in architecture, training data, and intended use cases. Each model generates a response under the same output format and length constraints. To assess the quality of these outputs, we employ a set of automatic evaluation metrics—previously used in the PersoBench benchmark~\cite{cho2022personalized}—that are specifically fine-tuned for evaluating coherence and personalization. The selected metrics offer more targeted insights than general-purpose measures like BERTScore or ROUGE. Moreover, they enable a structured comparative analysis of our framework against the referenced benchmark.
In addition to metric-aligned features for coherence and personalization, we include a length-format compliance signal to capture model instructability. Building on findings from PersoBench that highlight the influence of response failure rates on overall response evaluations, we place greater emphasis on instructability in our quality score computation. Preference pairs are then selected using margin-based sampling over the score distribution and used to fine-tune the model through a weighted preference optimization strategy.



\textbf{Preference Signals.} The quality of feedback used during language model tuning plays a critical role in shaping behavior~\cite{rafailov2023direct}. Building on this, we compute a quality score for each generated response by combining automated evaluation metrics capturing coherence, personalization, and instruction adherence, as introduced in PersoBench \cite{afzoon2024persobench}. Samples with clear score margins between accepted and rejected responses are retained, ensuring more focused and reliable fine-tuning.
To structure preference signals, we define two complementary types. The first targets response quality and includes four metric-based signals (UE, C, Coh-UniEval, and P), which emphasize coherence and persona alignment under the LLM-as-a-judge paradigm. Formally, the combined signal from this group is computed as:
Formally, the combined signal from this group is computed as:

\[
S_{\text{Metric-based}} = \frac{1}{|S|} \sum_{s \in S} s
\]

where \( S = \{\text{C Score}, \text{P Score}, \text{UE Score}, \text{Coh-UniEval Score}\} \) represents the set of metric-based evaluators.

The second signal type addresses instruction adherence, captured through a custom Length-Format Compliance (LFC) signal. LFC provides a bonus reward for outputs that satisfy the instruction constraints on format and response length, encouraging instruction compliance without rigid enforcement. Based on empirical observations, we assign higher weight to length compliance than to format compliance, as strict enforcement of format led to degraded generation quality. The signal is computed as:

\[
S_{\text{LFC}} = w_1 \cdot s_{\text{length}} + w_2 \cdot s_{\text{format}}, \quad \text{where } w_1 > w_2
\]

\textbf{Model Training.} Training pairs were constructed by computing a total quality score for each response and selecting pairs with a sufficient margin between chosen and rejected candidates, thereby reducing ambiguity and improving supervision reliability. Fine-tuning followed a preference-based objective inspired by DPO, where responses are ranked according to automatically computed quality signals:

\begin{equation}
\mathcal{L}_{\text{PersoDPO}} = 
\mathbb{E}_{x, y_c, y_r} \Big[ 
\sigma(\Delta S) \cdot \beta \cdot 
\big( \log p_\theta(y_c \mid x) - \log p_\theta(y_r \mid x) \big) 
\Big],
\end{equation}

where $\Delta S$ denotes the score margin between the chosen ($y_c$) and rejected ($y_r$) responses, and the total score $S_{\text{total}} = S_{\text{metric}} + S_{\text{LFC}}$. This objective encourages the model to assign higher likelihood to higher-scoring responses.





\section{Exprimental Setup}

\textbf{Dataset.} We used the FoCus dataset\footnote{\href{https://github.com/pkchat-focus/FoCus?tab=readme-ov-file}{https://github.com/pkchat-focus/FoCus}}, a benchmark for personalized response generation with multi-turn dialogues and persona annotations. Its longer contexts make it well-suited for testing coherence and personalization. Following PersoBench \cite{afzoon2024persobench}, prompts combined persona and dialogue context, with models instructed to generate a personalized response in a JSON format containing a single ``response'' field, ensuring consistency and enabling analysis of instruction adherence.

\textbf{Baselines.} We selected three open-source instruction-tuned LLMs—Qwen2-7B, Mistral-7B, and LLaMA 3.1-8B—to align with PersoBench \cite{afzoon2024persobench} and ensure consistent comparison. We also included a vanilla DPO-tuned Qwen2-5B-Instruct as a direct baseline for our custom DPO approach. Only models evaluated with dedicated alignment metrics were considered, discarding BLEU/ROUGE-based methods. The Qwen2-5B variant was chosen due to GPU memory constraints. All baselines were tested on 1,000 FoCus validation records with identical persona-augmented inputs and evaluation metrics.

\textbf{Evaluation Metrics.} Following PersoBench \cite{afzoon2024persobench}, we evaluated response quality with three automatic metrics covering personalization and coherence. Personalization was measured using the Consistency Score (C) and Persona Distance Score (P), while coherence was assessed with the Utterance Entailment Score (UE) and Coh-UniEval. All scores were normalized to [0,1].

\subsection{Implementation Setup}

\textbf{Baseline.} For initial response generation, we sampled \textbf{1,500 records} from the FoCus training set due to API query costs. In a zero-shot setup (\texttt{max\_tokens = 110}, \texttt{temperature = 0}), we prompted six LLMs: three open-source instruction-tuned models (\texttt{Qwen2-7B}, \texttt{Mistral-7B}, \texttt{LLaMA 3.1-8B}) and three API-based models (\texttt{GPT-3.5-Turbo}, \texttt{GPT-4-Turbo}, \texttt{GPT-4o-Mini}). The \texttt{max\_tokens} limit matched the observed upper bound of FoCus golden responses, ensuring comparable lengths. Deterministic generation was enforced by fixing \texttt{temperature = 0}. All models produced responses in a predefined JSON format with persona-augmented prompts. To construct the baseline DPO dataset, outputs were grouped by format validity: each training instance paired a prompt with a “chosen” (valid) and “rejected” (invalid) response.

\textbf{PersoDPO.} We created $\sim$9,000 preference pairs, each with a prompt, “chosen” and “rejected” responses, and quality scores. Final scores combined four metric-based features with an instruction-based feature assessing response length and JSON format compliance; the length component was weighted twice as much due to its stronger impact on generation stability \cite{afzoon2024persobench}. Fine-tuning was performed on \texttt{Qwen2-5B-Instruct} using a score-weighted DPO setup with the \texttt{TRL} library\footnote{\href{https://huggingface.co/docs/trl}{huggingface.co/docs/trl}} and the \texttt{ScoreWeightedDPOTrainer} class. Key hyperparameters included batch size = 4, gradient accumulation steps = 4, and 150 warm-up steps. The implementation is available upon request.

\section{Results and Discussion}



\begin{figure*}[!h]
\centering
\begin{subfigure}[b]{0.32\textwidth}
  \centering
  \includegraphics[width=\textwidth]{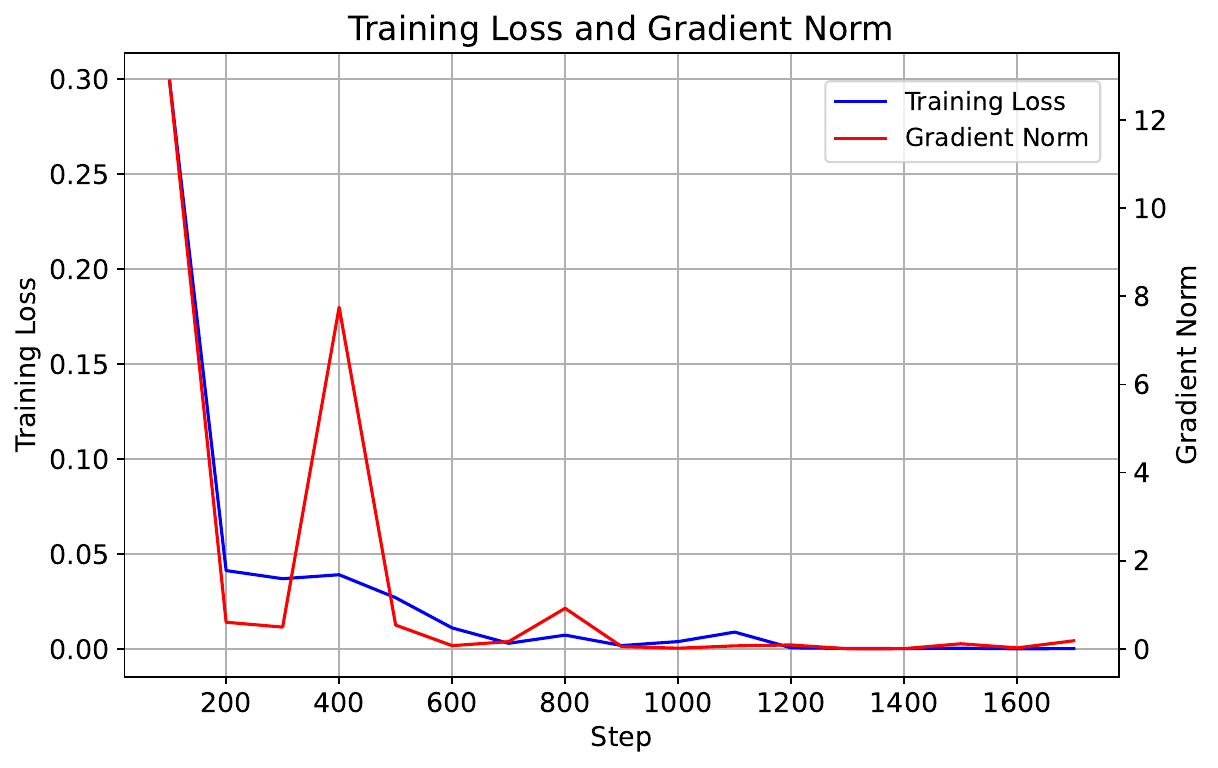}
  \caption{Loss Gradnorm}
  \label{fig:dpo-gradnorm}
\end{subfigure}
\hfill
\begin{subfigure}[b]{0.32\textwidth}
  \centering
  \includegraphics[width=\textwidth]{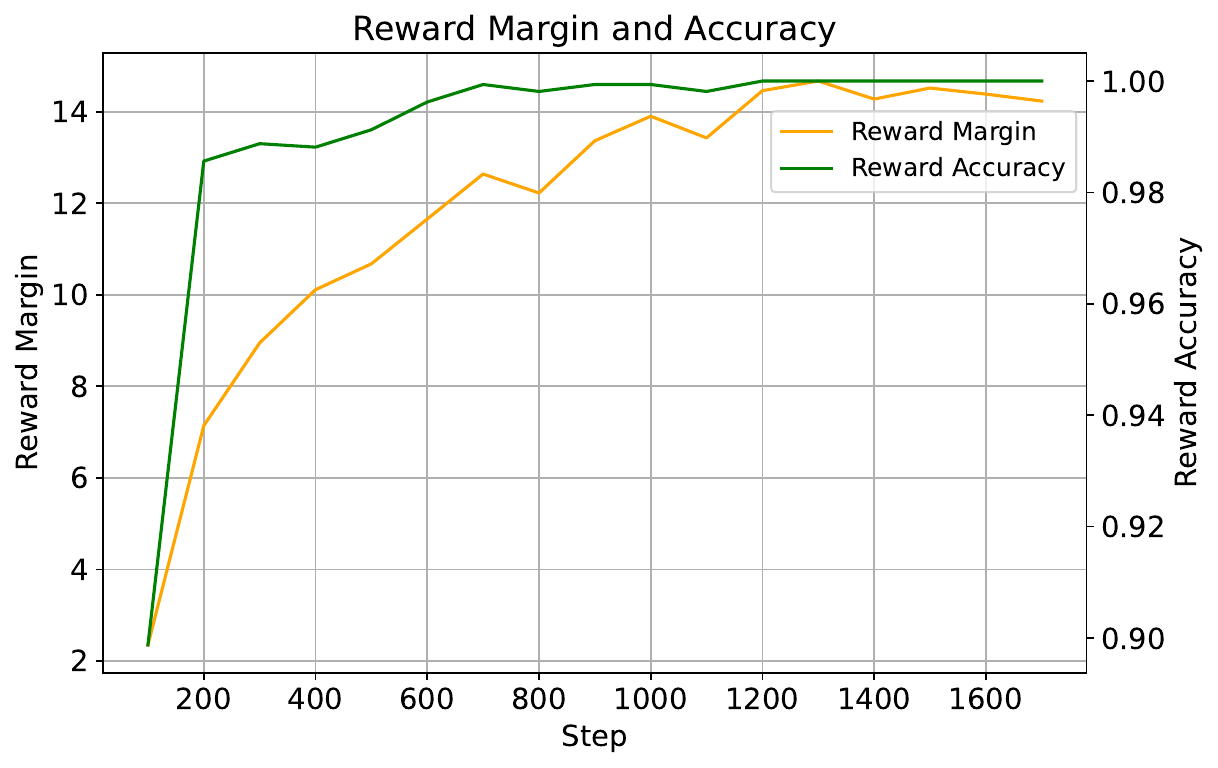}
  \caption{Margin-Accuracy}
  \label{fig:dpo-margin-accuracy}
\end{subfigure}
\hfill
\begin{subfigure}[b]{0.32\textwidth}
  \centering
  \includegraphics[width=\textwidth]{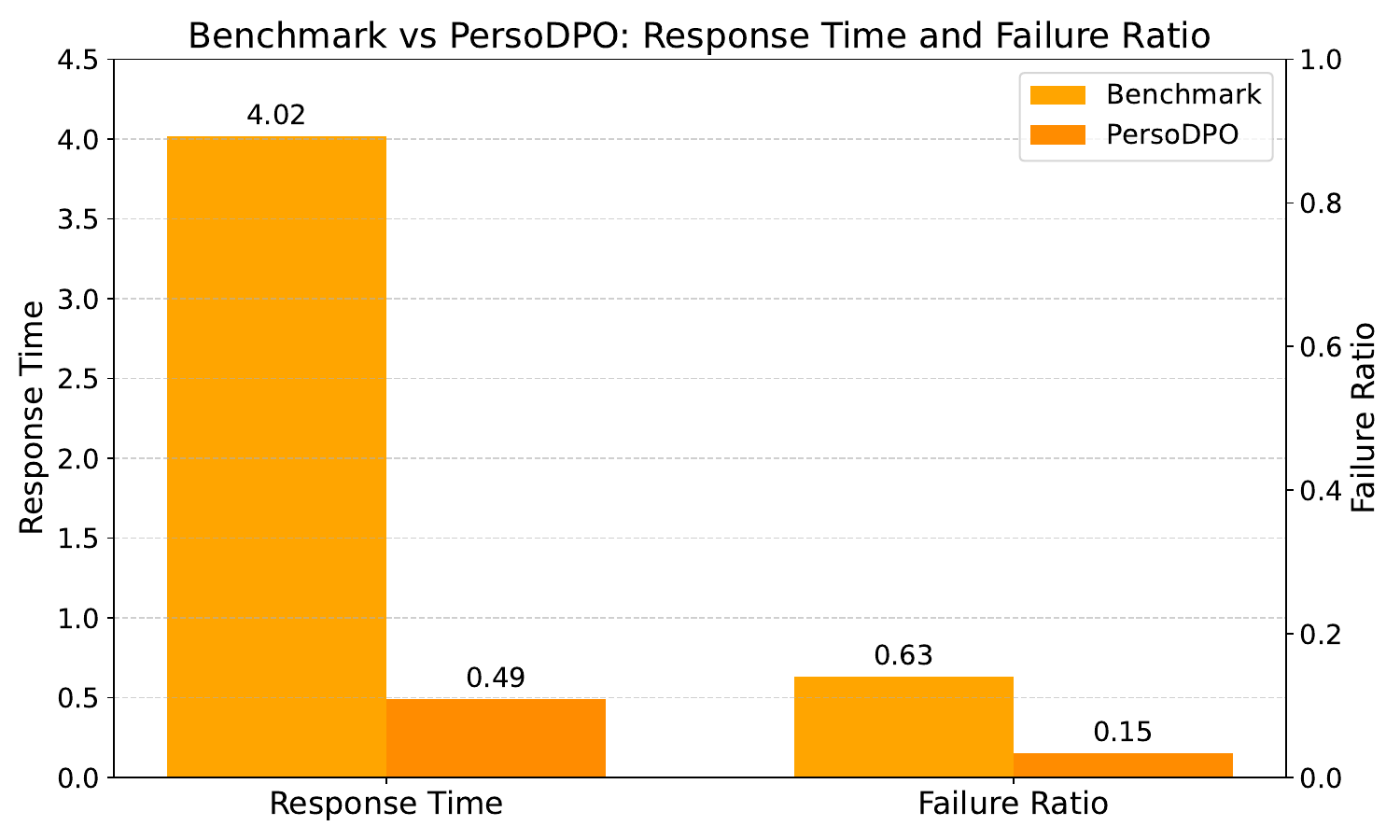}
  \caption{Response Time \& Failure Ratio}
  \label{fig:res-time-failure-ratio}
\end{subfigure}
\caption{Comparison of (a) Loss Gradnorm, (b) Margin-Augmented Accuracy, and (c) Response Time with Failure Ratio.}
\label{fig:combined-results}
\end{figure*}

Fig.~\ref{fig:combined-results} illustrates the optimization behavior of Qwen2-5B during PersoDPO fine-tuning. The training loss steadily declines with only minor early spikes in gradient norm, indicating smooth and robust convergence under score-guided supervision. Reward accuracy remains near the upper bound, while the reward margin narrows as the model learns to produce more consistent responses. Beyond training stability, PersoDPO achieves lower failure ratios—showing stronger adherence to length and format constraints—and faster response times, enhancing instructability, reliability, and responsiveness for industrial deployment.





\begin{table}[h!]
\caption{Result comparison of the proposed framework with the baselines.}
\centering 
\renewcommand{\arraystretch}{1.8}
\resizebox{0.7\columnwidth}{!}{%
\begin{tabular}{>{\raggedright\arraybackslash}p{4cm}cccc}
    \hline
    \textbf{Model}     &       \textbf{Coh-UniEval} & \textbf{C Score} & \textbf{UE Score} & \textbf{P Score} \\ \hline
    
    Qwen2 7B Benchmark &           0.37 $\pm$ 0.48     &  -0.31 $\pm$ 0.84  &  0.17 $\pm$ 0.48  &  0.29 $\pm$ 0.30 \\ \hline
    Mistral 7B Benchmark &         0.53 $\pm$ 0.46    &  -0.23 $\pm$ 0.82  &  0.21 $\pm$ 0.52  &  0.30 $\pm$ 0.28 \\ \hline
    Llama3.1 8B Benchmark &        0.55 $\pm$ 0.49     &  -0.22 $\pm$ 0.87  &  0.17 $\pm$ 0.48  &  0.33 $\pm$ 0.33 \\ \hline 
    Qwen2 5B DPO &                 0.98 $\pm$ 0.49     &  0.06 $\pm$ 0.49 &  0.21 $\pm$ 0.58 &  0.29 $\pm$ 0.26 \\ \hline\hline
    \textbf{Qwen2 5B PersoDPO} &   0.70 $\pm$ 0.45     &  \textbf{0.18 $\pm$ 0.87} &  \textbf{0.30 $\pm$ 0.62} &  \textbf{0.44 $\pm$ 0.30} \\ \hline\hline
    
\end{tabular}
}
\label{tab:result-comparison}
\end{table}

The performance of the proposed framework is evaluated using the metrics described in the previous section and compared against the baselines. As shown in Table~\ref{tab:result-comparison}, PersoDPO consistently outperforms all open-source benchmark models across all four evaluation metrics, demonstrating its strength in both response contextualization and personalization. This highlights the effectiveness of training with diverse LLM outputs, enabling PersoDPO to implicitly inherit complementary strengths from multiple pre-trained models. Considering the Coh-UniEval as one of the metrics designed to capture both contextual coherence and persona consistency, a remarkable improvement is observed, indicating PersoDPO’s strong capability in generating well-grounded and persona-aligned responses. Additionally, there is a tangible improvement in the UE Score as another metric evaluating the same aspects.
These gains suggest an enhanced ability to generate responses that are contextually appropriate while remaining aligned with persona traits. A consistent performance boost is also observed on C Score and P Score, which focus more directly on personalization, further confirming the benefit of multi-LLM preference modeling in capturing nuanced persona-driven behavior.
When compared to the vanilla DPO variant fine-tuned on Qwen2 5B, PersoDPO demonstrates clear improvements in three out of four evaluation metrics. It outperforms the baseline in both personalization-specific metrics—C Score and P Score—highlighting the effectiveness of multi-LLM preference modeling in capturing persona-aligned behavior. Notably, the P Score achieved by the vanilla DPO model not only fails to surpass any of the benchmarked LLMs but also matches the lowest among them, indicating its limited ability to model semantic alignment with persona traits. Although PersoDPO’s Coh-UniEval score is slightly lower than that of the DPO baseline, this is mitigated by its higher UE Score, which evaluates similar coherence and persona consistency dimensions. This balance suggests that PersoDPO maintains strong contextual grounding while offering significantly better personalization quality overall.
Among these metrics, C Score remains one of the most challenging for language models to perform well on, as it penalizes persona inconsistencies more severely than other evaluators, often assigning negative scores to misaligned responses. Nonetheless, PersoDPO achieves a substantial improvement over all baselines in this dimension, indicating its enhanced ability to maintain persona alignment even under stricter evaluative conditions. 
The raw responses generated by the proposed method, along with those from the baseline, are available in the GitHub repository for further analysis \footnote{\href{https://github.com/salehafzoon/PersoDPO/tree/main/Raw\%20Responses/FoCus}{github.com/salehafzoon/PersoDPO/tree/main/Raw\%20Responses/FoCus}}.



\section{Conclusion}
This study introduced PersoDPO, a preference optimization framework designed to improve both personalization and contextual coherence in persona-grounded dialogue systems. By leveraging outputs from diverse open- and closed-source LLMs, PersoDPO constructs high-quality preference pairs based on automatic evaluation signals without requiring manual annotation. The framework integrates coherence- and persona-driven metrics alongside a novel Length-Format Compliance feature, enabling more instructable and structurally consistent responses.
Empirical results on a personalized dialogue benchmark demonstrate that PersoDPO consistently outperforms strong open-source baselines and a vanilla DPO model across personalization and coherence metrics. Moreover, it shows clear advantages in industrially relevant aspects such as response time and instruction adherence, indicating its practical suitability for real-world deployment. 

\section*{Acknowledgments}
We acknowledge the Center for Applied Artificial Intelligence at Macquarie University~(Sydney, NSW, Australia) for supporting and funding this research.

\bibliographystyle{unsrt} 
\bibliography{references}

\end{document}